# Some Chances and Challenges in Applying Language Technologies to Historical Studies in Chinese

†Chao-Lin Liu ‡Guantao Jin §Qingfeng Liu †Wei-Yun Chiu ‡Yih-Soong Yu

†Department of Computer Science, National Chengchi University, Taiwan
‡National Chengchi University, Taiwan
§Institute of Chinese Studies, Chinese University of Hong Kong, Hong Kong
†Department of Chinese Literature, National Chengchi University, Taiwan
‡Department of History, National Chengchi University, Taiwan

†chaolin@nccu.edu.tw, ‡§gtqf1908@gmail.com

**Abstract.** We report applications of language technologies to analyzing historical documents in the Database for the Study of Modern Chinese Thoughts and Literature (DSMCTL). We studied two historical issues with the reported techniques: the conceptualization of "華人" and the attempt to institute constitutional monarchy in the late Qing dynasty. We also discuss research challenges for supporting sophisticated issues using our experience with DSMCTL, the Database of Government Officials of the Republic of China, and the *Dream of the Red Chamber*. Advanced techniques and tools for lexical, syntactic, semantic, and pragmatic processing of language information and more thorough data collection are in need to strengthen the collaboration between historians and computer scientists.

**Keywords:** Temporal analysis, keyword trends, collocation, Chinese historical documents, digital humanities, natural language processing, Chinese text analysis

## 1 Introduction

Natural language processing (NLP) is a well-known research area in computer science, and has been successfully applied to handle and analyze modern text material in the past decades. Whether we can extend the applications of current NLP techniques to historical Chinese text and in the humanistic context (e.g., Xiang and Unsworth, 2006; Hsiang, 2011a; Hsiang, 2011b; Yu, 2012) is a challenge. Word senses and grammars changed over time, when people assigned different meanings to the same symbols, phrases, and word patterns.

We explored the applications of NLP techniques to support the study of historical issues, based on the text material from three data sources. They include the Database for the Study of Modern Chinese Thoughts and Literature (**DSMCTL**, henceforth)[1], the Database of Government Officials of the Republic of China (**DGOROC**, henceforth)[2], and the *Dream of the Red Chamber* (**DRC**, henceforth)[3]. DSMCTL is a very large database that contains more than 120 million Chinese characters about Chinese history between 1830 and 1930. DGOROC includes government announcements starting from 1912 to present. DRC is a very famous Chinese novel that was composed in the Qing dynasty. These data sources offer great chances for researchers to study Chinese history and literature, and, due to the huge amount of contents, computing technologies are expected to provide meaningful help.

In this paper, we report how we employed NLP techniques to support historical studies. Chinese text did not contain punctuations until modern days, so we had to face not only the well-known Chinese segmentation problem but also the problem of missing sentence bounda-

---

1 中國近現代思想及文學史專業數據庫 (zhong1 guo2 jin4 xian4 dai4 si1 xiang3 ji2 wen2 xue2 shi3 zhuan1 ye4 shu4 ju4 ku4): http://dsmctl.nccu.edu.tw/d_about_e.html, a joint research project between the National Chengchi University (Taiwan) and the Chinese University of Hong Kong (Hong Kong), led by Guantao Jin and Qingfeng Liu

2 中華民國政府官職資料庫 (zhong1 hua2 min2 guo2 zheng4 fu3 guan1 zhi2 zi1 liao4 ku4): http://gpost.ssic.nccu.edu.tw/. The development of this database was led by Jyi-Shane Liu of the National Chengchi University.

3 紅樓夢 (hong1 lou2 meng4): http://en.wikipedia.org/wiki/Dream_of_the_Red_Chamber, a very famous Chinese novel that was composed in the eighteenth century ries. In recent attempts, we applied the PAT Tree method (Chien, 1999) to extract frequent Chinese strings from the corpora, and discovered that the distribution over the frequencies of these frequent strings conform the Zipf's law (Zipf, 1949).

We investigated the issue about how the Qing government attempted to convert herself from an imperial monarchy to a constitutional monarchy between 1905 and 1911, using the emperor's memorials (奏摺, zou4 zhe2)[4] about the preparation of constitutional monarchy[5]. To this end, we selected the keywords from the frequent strings with human inspection, and applied techniques of information retrieval to support the study.

We also studied the attitude of the Qing government towards the Chinese workers who worked in other countries between 1875 and 1911. We analyzed the co-occurrences, i.e. collocations, of the keywords over the years of interest, using the documents recorded in the diplomatic documents of the late Qing dynasty[6].

Detailed observations and discussions of these historical researches are reported in two other papers (Jin *et al.*, 2011; Jin *et al.*, 2012) that will be presented in the Third Conference of Digital Archives and Digital Humanities.

While we have applied NLP techniques to support historical studies, we have also experienced some challenging problems at the lexical, syntactic, semantic and pragmatic levels. For instance, what are the most appropriate computational functions that support a certain research need? Are the current databases good enough? We elaborate on these challenges based on our experience with the three data sources, i.e., DSMCTL, DGOROC, and DRC.

No one may expect that NLP techniques will replace the major roles of historians in the historical studies, but the techniques should be able to work with the historians to make the studies more efficiently and more effectively. Empirical experience reported in this paper and the literature have demonstrated the potential of NLP techniques. With the help of computing technologies, historians can delegate some search work and basic analysis to computers and spend more time on higher level philosophic issues than before.

## 2 Zipf's Law Applicability

The Database for the Study of Modern Chinese Thoughts and Literature contains six genres of text material that were published between 1830 and 1930. Except the first category, most of them were collected from the late Qing dynasty: modern periodicals, literati's personal publications, diplomatic documents, newspapers, official documents, and translated works by western commissioners. Currently, the database contains more than 120 million simplified Chinese characters.

For modern Chinese information processing with NLP techniques, researchers rely on good machine readable lexicons and good methods to segment Chinese strings into Chinese words. Both of these infrastructural facilities are missing for the processing of non-modern Chinese text. Hence, we bootstrapped our work by computing frequent Chinese strings with the PAT Tree technique in the documents, and asked historians to select relevant words from the frequent strings.

Table 1 shows the statistics about five collections in the DSMCTL database: *Constitution* (清末立憲檔案), *Diplomacy* (清季外交史料), *Min_Bow* (民報)[7], *Nations* (海國圖志)[8], and *New_People* (新民叢報)[9]. They contain about 11 million characters, about one tenth of the whole DSMCTL database. We refer to strings that occurred more than 10 times[10] in a collection as ***pseudowords***. Many of these pseudowords have specific meanings, but not all of them do.

[4] Most Chinese words are followed by their Hanyu pinyin and tone when they appear the first time in this paper.

[5] 清末籌備立憲檔案史料 (qing1 mo4 chou2 bei4 li4 xian4 dang3 an4 shi3 liao4) : http://baike.baidu.com/view/3299810.htm

[6] 清季外交史料 (qing1 ji4 wai4 jiao1 shi3 liao4): http://zh.wikisource.org/zh-hant/清季外交史料選輯

[7] 民報 (min2 bao4): http://zh.wikipedia.org/wiki/民報

[8] 海國圖志 (hai3 guo2 tu2 zhi4): http://zh.wikipedia.org/wiki/海國圖志

[9] 新民叢報 (sin1 min2 cong2 bao4): http://zh.wikipedia.org/wiki/新民叢報

We ranked the pseudowords based on their frequencies, i.e., the most and the second most frequent pseudowords were ranked first and second, respectively. Then, we computed the logarithmic values of the ranks and frequencies, and drew the curves in Figure 1. The curves in Figure 1 indicate that the pseudowords in the Chinese historical documents, like documents written in modern English and Chinese languages (Ha *et al*., 2003; Xiao, 2008), conform to the Zipf's law quite well (Zipf, 1949).

Let $r$ and $f$ denote the rank and frequency of a word in a collection of text, respectively, Zipf's law predicts that the product of $r$ and $f$ is a constant, $c$, as is shown in Equation (1).

$$f = \frac{c}{r} \tag{1}$$

Hence, we will observe curves that are almost straight lines after we take the logarithm (usually abbreviated as "log") on both sides of Equation (1) to become Equation (2). In Figure 1, the log values of pseudoword frequencies are on the vertical axis, and the log values of the pseudoword ranks are on the horizontal axis.

$$\log(f) = \log(c) - \log(r) \tag{2}$$

Let $N$ denote the total number of characters in a collection. We divided the word frequencies by the sizes of individual collections. In Figure 2, the vertical axis shows the log values of the pseudoword frequencies divided by $N$, namely $\log\left(\frac{f}{N}\right)$. The curves for the distributions of the pseudowords almost overlap, suggesting that the Zipf's law applied to the five collections quite uniformly, after we considered the influences of the sizes of collections.

The decision to divide term frequency, $f$, by the corpus size, $N$, was arbitrary, but it was very interesting to find that curves in Figure 2 almost overlap as a result. Evidently, sizes of corpora affected the shapes and positions of the Zipfian curves. Xiao (2009) attempted to study the influences of corpus size over the Zipfian curves. In one of the reported study, Xiao sampled five small datasets of almost the same size from the General Contemporary Chinese Corpus, which contained approximately one billion Chinese characters. Zipfian curves drawn for these datasets almost overlapped perfectly.

## 3 Chronicle Trends of Multiple Keywords

We can examine the pseudowords and selected those that are potentially relevant to the historical issues as ***keywords***. We computed the annual and total frequencies of each of these keywords, and computed the total number of keywords in each year.

The "Total" curve serves as the basis for the analysis of importance of keywords. Let $t_{1905}$, $t_{1906}$, $t_{1907}$, $t_{1908}$, $t_{1909}$, $t_{1910}$, and $t_{1911}$ denote the total number of keywords appeared in 1905, 1906, 1907, 1908, 1909, 1910, and 1911, respectively. We could compute the total number of keywords in *Constitution*, $T$, using the following equation.

$$T = t_{1905} + t_{1906} + t_{1907} + t_{1908} + t_{1909} + t_{1910} + t_{1911} \tag{3}$$

Using the years on the horizontal axis and the ***annual percentage***, $\frac{t_i}{T}$, on the vertical axis, we analyzed the keywords in *Constitution* (cf. Table 1) to obtain the "Total" curve in Figure 3.

Analogously, let $K$ denote the total number of a particular keyword, e.g., "官制"[11], that appeared in *Constitution* and $k_n$ denote the number of all of the keyword appeared in a year $n$.

---

[10] The selection of 10 as the threshold was by the historians. The choice was heuristic but arbitrary.

[11] 官制: guan1 zhi4, bureaucracy

We can draw a curve of annual percentage for a keyword. Figure 3 shows the curves of annual percentages of all words (Total) and six keywords over the years between 1905 and 1911 in *Constitution* (cf. Table 1).

When the keywords appeared more frequently in a year, a historical event typically coincided (Jin *et al*., 2011). We considered the keyword to be special in the year *n*, if $\frac{k_n}{K} \geq \lambda \frac{t_n}{T}$. In (Jin *et al*., 2011), we chose $\lambda = 1.1$ arbitrarily, but the selection of $\lambda$ can be adjusted as needed in a computer-assisted document analysis environment.

For instance, in 1906, $\frac{k_{1906}}{K}$ for "官制" was about 0.45, and $\frac{t_{1906}}{T}$ was less than 0.25. In 1907, $\frac{k_{1907}}{K}$ for "立宪"[12] was about 0.40, and $\frac{t_{1907}}{T}$ was less than 0.33. Both "官制" and "立宪" qualified as special. In 1906 and 1907, the Qing government began to consider constitutional monarchy seriously, so government officers discussed the issues about "bureaucracy" ("官制") and "constitutionality" ("立宪") for running the new form of government intensively. Hence, the keywords like "官制" and "立宪" appeared in the emperor's memorials relatively often than other years.

In 1908, $\frac{k_{1908}}{K}$ for "选举"[13] was about 0.45, and $\frac{t_{1908}}{T}$ was less than 0.15. In fact, in 1908, the keywords about election ("选举" and "章程") were used more frequently in the emperor's memorials.

After years of discussion about the fundamental issues about a constitutional monarchy, the Qing government appeared to be prepared for the new form of government, and was taking steps for its realization. In 1909 and 1910, words about self governance ("筹办" and "自治")[14] were becoming relatively more important.

The temporal relationship between these six keywords' emerging importance further suggested the progression toward the establishment of a constitutional monarchy before the overthrown of emperor in late 1911. Namely, the focus of discussion shifted from planning and preparation to realization and actions.

Our approach is more appropriate for historical studies than the Google Trends[15], although the difference is subtle and may appear minuscule. The analysis of occurrence of an individual keyword, like the Google Trends, is useful for studying the changing importance of a keyword over a period of time. Evaluating the chronicle change of importance of a keyword is certainly important, but we further compare the chronicle changes of multiple keywords, which allows us to visualize the trends more directly.

## 4 Temporal Analysis of Important Collocations

A ***collocation*** is formed by two keywords that appeared "close" to each other in a statement. A collocation carries more specific semantic information than an individual keyword. The occurrence of the keyword "Chinese labor" ("华工", hua2 gong1) could have referred to anything about Chinese labor, e.g., limiting ("限制", xian4 zhi4) or protecting ("保护", bao3 hu4) the Chinese labor, while a collocation "protect the Chinese labor" ("保护" and "华工") provides more specific meaning than the individual keywords.

However, given that there were neither word boundaries nor sentence boundaries in ancient Chinese documents. We chose to define "close" based only on the "distance" between two keywords.

---

[12] 立宪: li4 xian4, constitutionality
[13] 选举: xuan3 ju3, election; 章程: zhang1 cheng2, rules
[14] 筹备: chou2 bei4, preparation; 自治: zi4 zhi4, self governance
[15] http://www.google.com/trends/

A keyword is considered to be collocated with another if the keywords were less than 30 characters apart. Our computer programs were flexible in setting the window size for "closeness". We defined *collocation window* as the span of characters around a keyword that are considered "close". We ran experiments in which sizes of the collocation windows were set to 10, 20, and 30 characters. A collocation window of 30 characters will consider 30 characters on the left and on the right side of a keyword, for instance. The historians observed the computed collocations and preferred the size of 30.

We analyzed the statistics of collocations in the documents about the concept of "Chinese People" ("华人", hua2 ren2) in *Diplomacy* (cf. Table 1). We identified the keywords with the procedure that we applied to find individual keywords in *Constitution* that we explained in the previous section. Historians then chose the keywords of interest and we ran the computer programs to do the temporal analysis of the important collocations. This procedure is similar to the procedure that we used to obtain Figure 3; the only difference was whether the target of analysis was keywords or collocations.

Curves in Figure 4 show that the annual percentages of four collocations varied over the years between 1875 and 1909. Significantly large annual percentages coincided with the historical events of the years, again. In 1894, the United States of America (USA) ("美国"(mei3 guo2) in the chart) and the Qing government signed a treaty to limit Chinese labor("华工") to enter USA.[16] In 1905, Chinese societies started to boycott American's products mostly because the USA would extend the treaty signed in 1894.[17] Initially, the Qing dynasty was trying to protect only the Chinese labor. Later, the protection extended to *Chinese merchants* and then extended to *Chinese people* (Jin *et al.*, 2012).

## 5 Ranking Individual Documents: An Application of Information Retrieval

As the statistics in Table 1 showed, there can be more than thousands of documents that contain millions of characters in a particular collection. Finding the most relevant documents or essays to read is not easy in the past. With our ability to identify the important keywords and collocations, we could rank the documents based on how documents included the important keywords and collocations. Table 2 shows a part of the table in which we ranked the documents in *Constitution* (cf. Table 1). The weights in Table 2 were calculated based on the number of keywords that were used in a document. Larger weights imply that more keywords were used in the document, so the document might be more relevant to the research topic for which the researcher selected the keywords. The ranking function and other techniques for information retrieval and extraction could provide useful information for the historians to study specific issues (Jin *et al.*, 2011; Jin *et al.*, 2012).

## 6 Discussions

In this section, we discuss some technical problems that are related to using computing techniques to support historical studies in Chinese.

### 6.1 Lexical Ambiguity, Pragmatics, and Term Identity

We have illustrated three possible applications of text analysis for historical studies in previous sections. The applications were based on the frequencies of keywords in the text collections. In NLP, we can refer to the frequencies of keywords as ***term frequencies***. In addition, we relied on the "time stamps" of the documents, where the "time stamps" are the recorded times of the documents. Based on our dependence on the terms frequencies and time stamps we obtained and presented the figures that we discussed in Sections 3 and 4.

---

[16] 《中美華工條約》、《限禁來美華工保護寓美華人條約》：http://dict.zwbk.org/zh-tw/Word_Show/64744.aspx
[17] http://zh.wikipedia.org/wiki/抵制美貨運動；《籌拒美國華工禁約公啓》

In these examples, we presume that the frequencies reflect the importance of the concepts that are represented by the terms and collocations, and results of our work are quite convincing. However, we have to watch for the problems of lexical ambiguity and pragmatics that are hidden under the term frequencies.

For instance, frequently cited events of the past may induce confusion about the significance of term frequencies. Tu *et al.* (2011) discover that, although “張公藝” (zhang1 going1 yi4) was a Chinese name that appeared frequently in collections in the Taiwan History Digital Library (THDL). “張公藝” referred to a person who actually lived in the Tang dynasty (618AD-1907AD)[18], which is well before the time period of the documents in THDL. The documents in THDL referred to“張公藝” because of a story that was well-known in the Qing dynasty (1644AD-1912AD). That the term frequency of “張公藝” is high in THDL does not imply that “張公藝” himself was an important person in Taiwan in Qing dynasty.

Lexical ambiguity may make the term frequencies less reliable. Yu (2012) accentuates this issue with “民主” (min2 zhu3). In modern Chinese, “民主” is the word for “democracy”. However, it could represent the emperor (民之主, min2 zhi1 zhu3), the American president, and the Republic (in 民主國, min2 zhu3 guo2) in non-modern Chinese text.

The first author of this paper examined the DGOROC, and found that “陳建中” was a very common names in the database. Hence, finding the actual identities of names is an important issue, in addition to computing the term frequencies. Distinguishing persons of the same names in modern databases requires extra-ordinary sources of private information.

Although differentiating persons with the same names is not easy, identifying names in Chinese text is not an easier task for the research of Named Entity Recognition (often referred as NER, e.g., Wu *et al.*, 2006) in the first place. For instance, it may not be easy to extract names from Chinese text like “中央高層正醞釀安排令計畫接任中組部長”[19] (zhong1 yang1 gao1 ceng2 zheng4 yun4 niang4 an1 pai2 ling4 ji4 hua4 jie1 ren4 zhong1 zu3 bu4 zhang3) if we did not know “令計劃” (ling4 ji4 hua4) is a name.[20]

### 6.2 Word Segmentation and Sentence Division

In Section 6.1, we discuss that the interpretation of a given term. In fact, we have to define the concept of “term” in Chinese. If we cannot define terms precisely, then we would have no ground for defining collocations. It is well known that Chinese words are not separated by spaces like alphabetic languages. The task of separating Chinese words in Chinese text is generally called word segmentation (e.g., Ma and Chen, 2005; Jiang *et al.*, 2006; Tseng *et al.*, 2005). In contrast, it is less known that ancient Chinese text does not have punctuations, and readers have to figure out the divisions of sentences (Huang, 2008).

Clearly if we could not divide sentences and segment words correctly, we would not be able to acquire correct term frequencies. This may happen when we process text like “五行者金主義木主仁水主智火主禮土主信” (wu3 xing2 zhe3 jin1 zhu3 yi4 mu4 zhu3 ren2 shui3 zhu3 zhi4 huo3 zhu3 li3 tu3 zhu3 xin4). We would have to add punctuations to divide this string: “五行者，金主義，木主仁，水主智，火主禮，土主信”. After this step, we know that “智火” is not a term in the original string, although “智火” can be a meaningful term in modern Chinese.[21] Given the divided string, we still have to face the word segmentation problem. In this example, each character in “金主義” represents a specific meaning. We cannot interpret “主義” in “金主義” as we would interpret “主義” (-ism) in “帝國主義” (di4 guo2 zhu3 yi4; imperialism) or “資本主義” (zi1 ben3 zhu3 yi4; capitalism) in modern Chinese. Similarly, we have to know that “金主” is not a term in the original string, although “金主” (a wealthy person) is a meaningful term in modern Chinese.

[18] http://en.wikipedia.org/wiki/Tang_Dynasty

[19] Source: http://www.cbfcn.com/news_detail.aspx?strnew=1154

[20] In fact, “令路線” (ling4 lu4 xian4), “令政策” (ling4 zheng4 ce4), “令完成”(ling4 wan2 cheng2), and “令方針”(ling4 fang1 jhen1) are also Chinese names (although they are quite unusual) : http://zh.wikipedia.org/wiki/令計劃

An actual problem took place when we used the DSMCTL to investigate whether energy conservation was a concern in the Qing dynasty (Chou *et al.*, 2012). Without a Chinese segmenter for ancient Chinese, we found many occurrences of “能源” (neng2 yuan2) in the database, but, most of the time, “能源” was just a sub-string of “不能源源而來” (bu4 neng2 yuan2 yuan2 er2 lai2) when people talked about something that could not come indefinitely.[22]

### 6.3 Trends: Informative or Deceptive

In Sections 3 and 4, we briefly introduced applications of temporal trends of keywords (Figure 3) and trends of collocations (Figure 4), that were more thoroughly discussed in (Jin *et al.*, 2011) and (Jin *et al.*, 2012), respectively. Researchers in other fields also found impressive applications of trends of keywords (e.g., Caneior and Mylonakis, 2009). Despite these successful applications, caution is in need to interpret the observed trends.

Figure 5 shows temporal trends for the names of three main characters in a famous novel *Dream of the Red Chamber* (DRC). The horizontal axis shows the chapters of the DRC. The vertical axis shows the frequency of the keywords (persons' names in this chart). The highs of the curves shows the times of being mentioned of a person in a particular chapter, so are indicative of the relatively importance of the persons. We discuss three main persons in DRC, “寶玉” (Bao3 Yu4), “黛玉” (Dai4 Yu4), and “寶釵” (Bao3 Tsai1), in the following.

Do the ups and downs of a particular curve show the changes of importance of a person? Intuitively, the answer may be yes. If the name of a person was mentioned more frequently, that particular person should be more involved in a chapter. However, this interpretation is not flawless – A person was mentioned more times might be a result of a longer chapter. Being mentioned more times in a longer chapter might not be a good proof for the importance of the mentioned person.

Figure 6 shows the numbers of characters in each chapter in DRC. Evidently, some chapters are longer and some are shorter.

Let $f_t$ and $l_t$ , respectively, denote the frequency of a keyword and the length of a chapter $t$ in DRC. We divide $f_t$ by $l_t$, for $t$=1, …, 120, for the three names in Figure 5. Figure 7 shows the resulting curves for the three persons.

We can observe some important changes in the curves. Take the curve for Bao-Yu (“寶玉”) for example. Bao-Yu was mentioned 84, 116, and 98 times in Chapters 8, 19, and 28, respectively. These three instances formed the first three peaks above 80 in the curve for Bao-Yu in Figure 5. The frequencies may have suggested that Bao-Yu were more important in Chapter 19 than in Chapters 8 and 28. After we divided these frequencies by the chapter lengths, we observed that the proportions of Bao-Yu being mentioned in these chapters were almost the same in Figure 7. Hence, the trends illustrated in Figures 6 and 7 provide hints for different conclusions.

---

[21] “智火” happens to be the name of a Chinese company: http://www.zhihuo.asia/.

[22] The Chinese segmentation service at Academia Sinica (http://ckipsvr.iis.sinica.edu.tw/) would return “不能”, “源源”, “而”, and “來” for “不能源源而來”. The online version of the Stanford parser (http://nlp.stanford.edu:8080/parser/index.jsp) would return “不”, “能”, and “源源而來”.

Consider another example. Assume that we want to know who among the three persons liked to “smile and say” most in DRC. Curves in Figure 8 show the frequencies of “寶玉笑道”, “黛玉笑道”, and “寶釵笑道” in DRC, where “笑道” (xiao4 dao4) is a way to say “smile and say”. The curves suggest that, before Chapter 40, Bao-Yu was the person who liked to “smile and say” most.

However, one may contend that the absolute frequency may not be a perfect indicator for how likely a person was to “smile and say”. If a person was mentioned less frequently, than s/he would not be able to “smile and say” as frequently as another who was mentioned relatively more frequently.

Let $s_t$ and $m_t$, respectively, denote the frequency a person “smiled and said” and a person was mentioned in a chapter $t$. For the three persons in our current discussion, $m_t$ was their individual term frequency $f_t$ that we showed in Figure 5. We divided $s_t$ by $m_t$ for each person and came up with Figure 9.

Quite interestingly, the curve for Bow-Yu does not dominate the others anymore. Instead, Bao-Tsai (“寶釵”) smiled and said something once every two appearance in Chapter 19, so did Dai-Yu (“黛玉”) in Chapter 73. In fact, never did Bao-Yu smile and say as often as 50% of the time he appeared in any chapter. The highest proportion of Bao-Yu’s “smile and say” took place in Chapter 88, where the proportion still fell short of 40%.

A researcher may be interested in the times a person smiled and said something, while another might be interested in the proportion a person smiled and said something when the person was mentioned in DRC. Take Bao-Yu for example. In the former case (Figure 8), the term frequency of Bao-Yu is the focus. In contrast, in the latter case (Figure 9), the conditional probability $\Pr(\text{Bao-Yu-Xiao-Dao} \mid \text{Bao-Yu})$ is of interest, and we have to compute the probability based on the observed frequencies. Different trends and analyses should be used for different purposes, and this is up to the researchers’ discretion. While designing tools for assisting historical studies, appropriate functionalities should be considered and explained to their users as clear as possible.

### 6.4 Transliteration and Translation

In addition to the ability to process normal ancient Chinese text, one may need to handle transliterated and translated words. Chinese people encountered western culture more directly and more frequently starting from late 1500s. Transliteration and translation are important ways for people to use Chinese words to convey and understand western concepts and entities.

To study the interactions between the Chinese and western cultures in ancient times, getting to know the Chinese transliterations and translations is an important step. For instance, “president” was transliterated into “伯理璽”(bo2 li3 si2), “伯理喜頓”(bo2 li3 si2 dun4), “伯理璽天德”(bo2 li3 si2 tian1 de2), and “伯力錫天德”(bo2 li3 si2 tian1 de2). “Pacific Ocean” was translated into “大海”(da4 hai3), “大東洋”(da4 dong1 yang1), and “太洋海”(tai4 yang1 hai3), and transliterated into “卑西溢湖”(bei4 si1 yi1 hu2) and “比西非益海”(bi3 si1 fei1 yi4 hai3). “Politics” was transliterated into “薄利第加”(bo2 li4 di4 jia1) and “波立特”(po2 li4 te4).

Historians may spend their lifetimes to identify the translated and transliterated terms in historical documents. If one could provide researchers the Chinese terms for the western concepts and entities, the researchers would be able to investigate and understand how Chinese faced the West hundreds years ago.

Therefore, we imagine that it would be terrific if computing technologies can help historical researchers identify transliterated and translated terms in historical documents. It may be not easy to use human experts to annotate a database that has 120 million characters in DSMCTL.

### 6.5 Advanced NLP Techniques: Trend Analysis

An anonymous reviewer of the manuscript of this paper points out that applications of advanced NLP techniques will strengthen the values of the collected statistics. For instance, one may classify the keywords into types, and conduct temporal analysis on keywords of the same type. This may give us a trend analysis similar to the analysis of emotion trend reported in (Yang *et al.*, 2007). It is also possible to treat the network of keywords as a social network. The nodes can be verbs, nouns, and names, and links show strengths of associativity. Networks like this may shed light on historical events that were hard to be revealed by simply studying the historical documents.

### 6.6 Time Stamps, Missing Data, and Fundamental Changes in DGOROC

The DGOROC database[23] provides information about the appointments of government officials of the Republic of China in Taiwan. This database was constructed and verified with human labor. Information was copied from hard copies of official documents, entered into text files, and was verified for quality assurance. It contains more than 850 thousand records dated from 1912, and is useful for studying modern history and relevant applications about Taiwan and China (e.g., Liu and Lai, 2011).

Since the data came from a real and changing government, there can be barriers that were hard to overcome simply by computing technologies. For instance, the current government in Taiwan was not in a really stable condition until she moved to Taiwan in 1949. Hence, the database is relatively more complete for records after circa 1949.

It should not be surprising that a government tries and evolves to serve the nation in the fast changing world. For instance, there was no "Ministry of Education" before 1928, although there must have been some government agents to handle national education policies before 1928. Hence, a researcher will have to know the names of the agents that were responsible for education to study the national education policies circa 1928. In this case, a simple keyword search service may not help very much.

Although the data collected after 1949 was more complete in DGOROC, the government may change the rules for whether or not to announce some types of assignments. For instance, there are departments in the Ministry of Education, and the department heads may change their appointments from a department to another, but this type of switch is not publicly announced in recent years.

The appointments of lower ranks of government officials may not be announced at real time. The announcement of such appointments may be delayed so that a larger group of appointments would be announced at the same time. If the time stamps of events for a certain study matter, then this kind of delay may be troublesome.

Despite these remaining challenges in DGOROC, we consider this database unique and important. By incorporating information available from other database maintenance agents of the central government and from national libraries, the database will offer researchers a great information source for studying modern history of Taiwan.

[23] The first author gained experience with DGOROC while serving as the project leader for maintaining DGOROC between February and August 2011. The comments about DGOROC in this section are of the first author's.

## 7 Concluding Remarks

We delineate our experience in using three sources of historical documents in Chinese: the database of Chinese historical documents that contain more than 120 million simplified Chinese characters, the Database of Government Officials of the Republic of China, and the Dream of the Red Chamber. Techniques for natural language processing were employed to analyze the contents of the documents to facilitate the studies of historical events. The exploration showed that NLP techniques are instrumental for the studies of non-modern Chinese historical documents. Our experience also suggested that advanced NLP techniques and more complete data collection are necessary for supporting research work in more precise ways.

## Acknowledgments

The work was supported in part by the funding from the National Science Council in Taiwan under the contracts NSC-99-2221-E-004-007 and NSC-100-2221-E-004-014. The Database for the Study of Modern Chinese Thought and Literature (1830-1930) was funded by the subproject 100H51 at the National Chengchi University, Aim for the Top Universities Project, Ministry of Education, Taiwan. The writing of this paper benefits tremendously from reviewers' comments on previous versions of this paper. Section 6.5, in particular, was prompted by the review comments of this journal version.

**Table 1. Statistics for five collections**

| Collection | Number of Different Pseudowords | Total Number of Characters | Number of Different Characters | Number of Documents |
|---|---|---|---|---|
| Constitution | 3288 | 713131 | 4097 | 399 |
| Diplomacy | 29315 | 2875032 | 5225 | 5758 |
| Min_Bow | 7784 | 1450623 | 6230 | 325 |
| Foreigns | 2649 | 679410 | 4916 | 160 |
| New_People | 33378 | 5259590 | 6647 | 1524 |

**Table 2. Three highly ranked emperor's memorials written in 1906 (in *Constitution* in Table 1) and their weights (i.e., relative importance), authors, and titles**

| **1906** | | |
|---|---|---|
| **Weights** | **Authors** | **Document Title** |
| 420 | 戴鸿慈 | 出使各国考察政治大臣戴鸿慈等奏请改定全国官制以为立宪预备折 |
| 312 | 杨晟 | 出使德国大臣杨晟条陈官制大纲折 |
| 122 | 殷济 | 内阁校签中书殷济为豫备立宪条陈筹经费建海军等二十四条呈 |

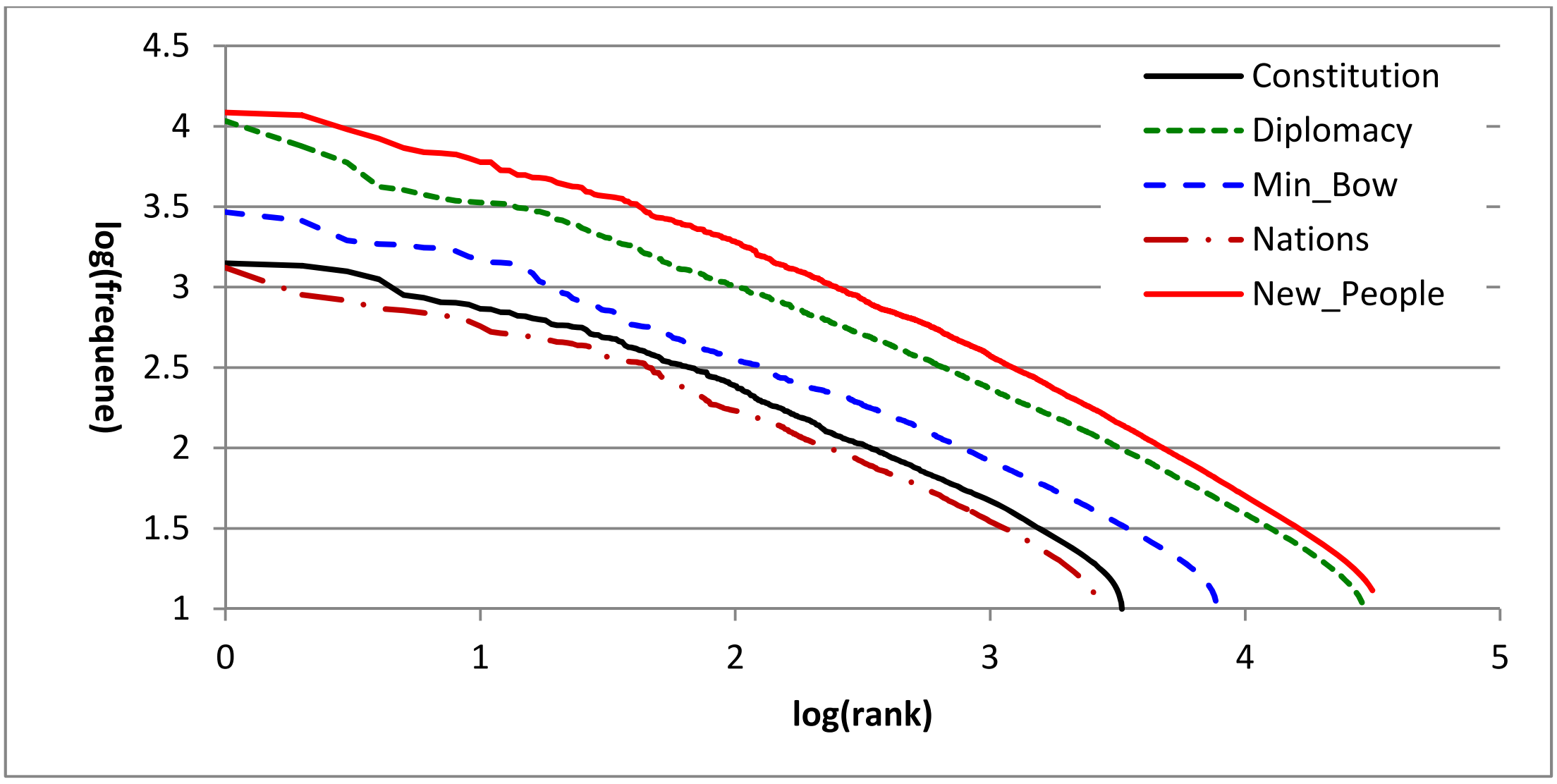

**Figure 1. Pseudowords in the Chinese historical collections abide by the Zipf's law**

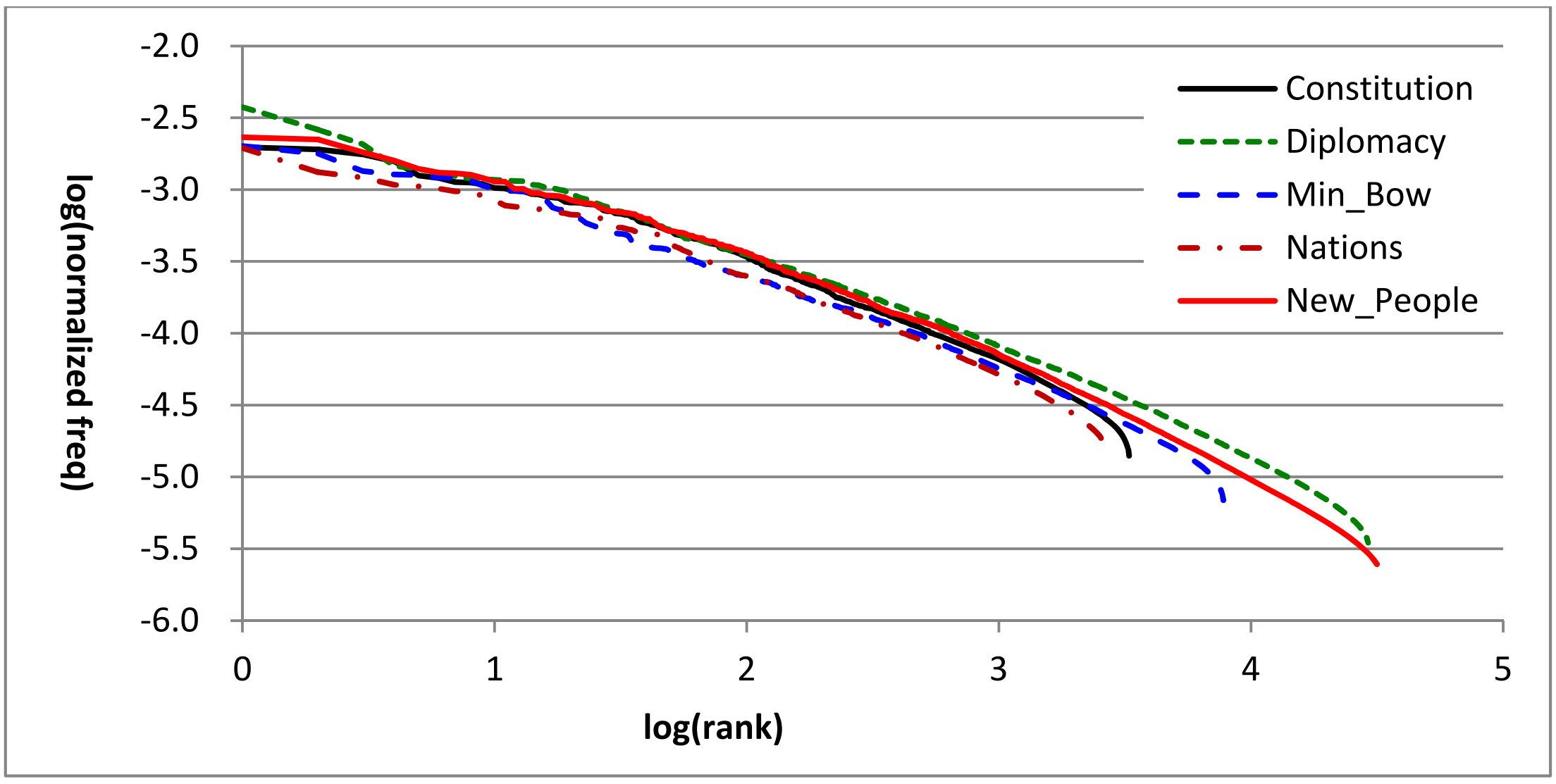

**Figure 2. Reducing the influences of sizes of individual collections**

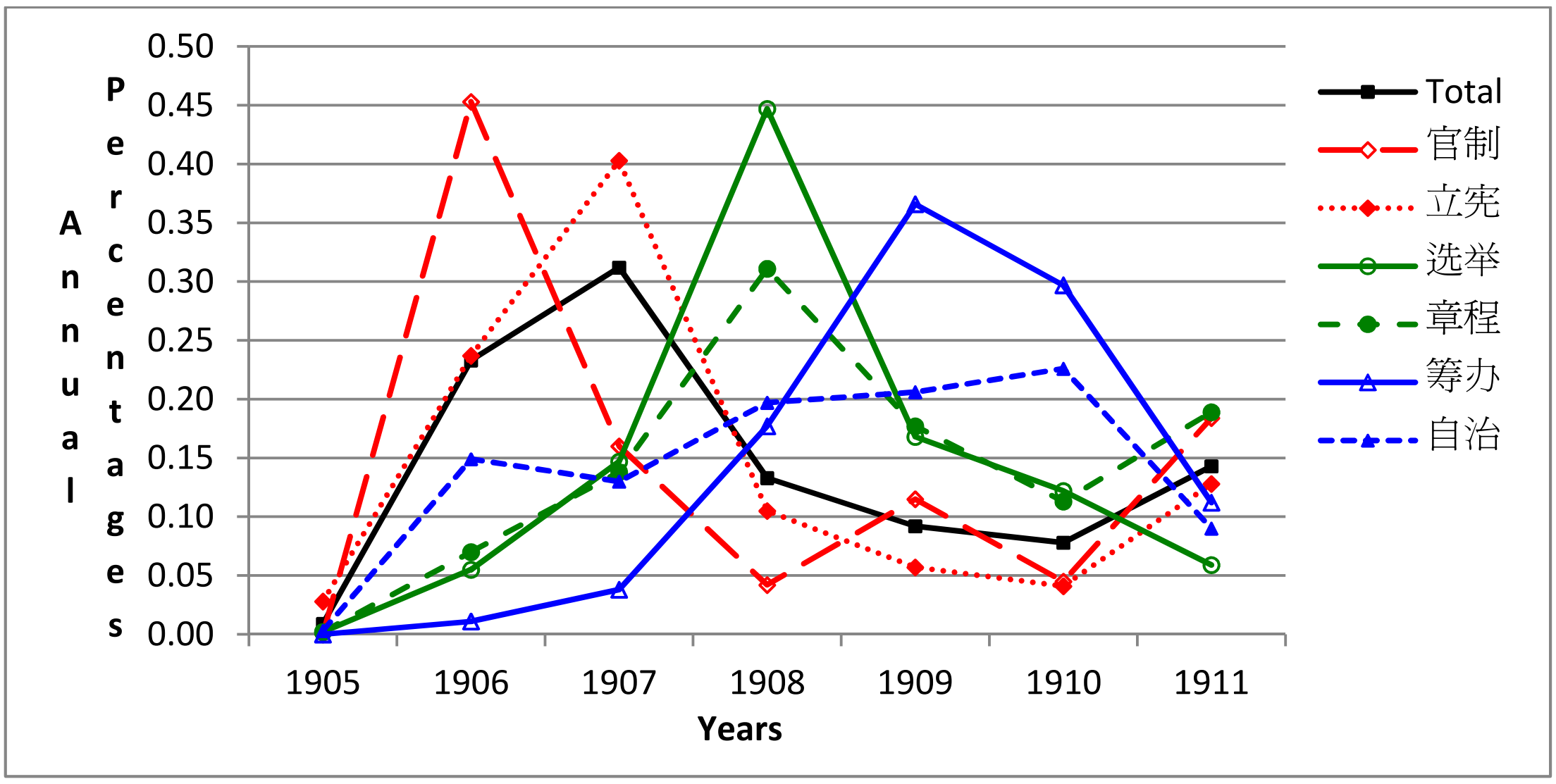

**Figure 3. Some keywords appeared more frequently in particularly years**

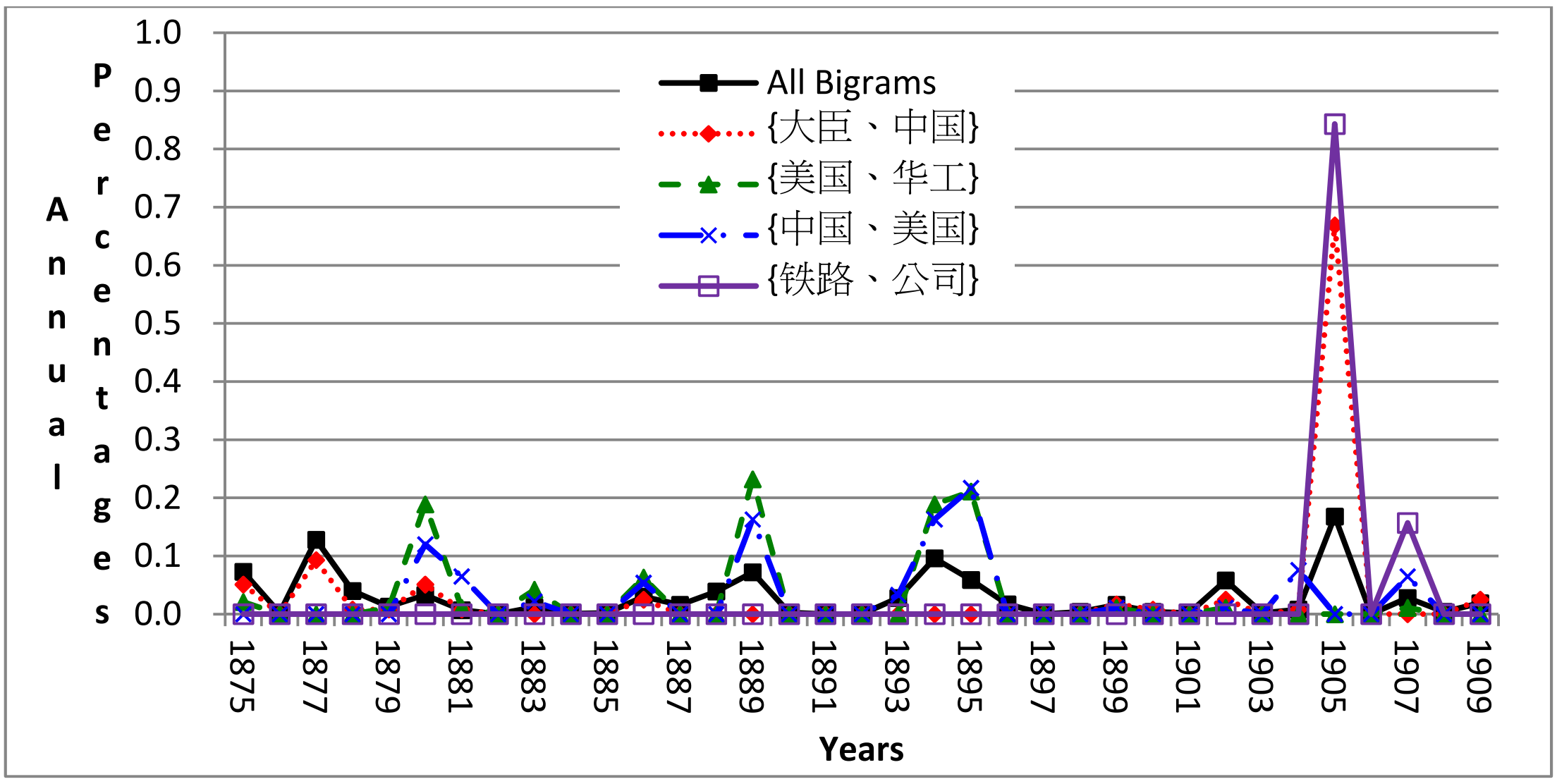

**Figure 4. Importance of keyword collocations varied over the years**

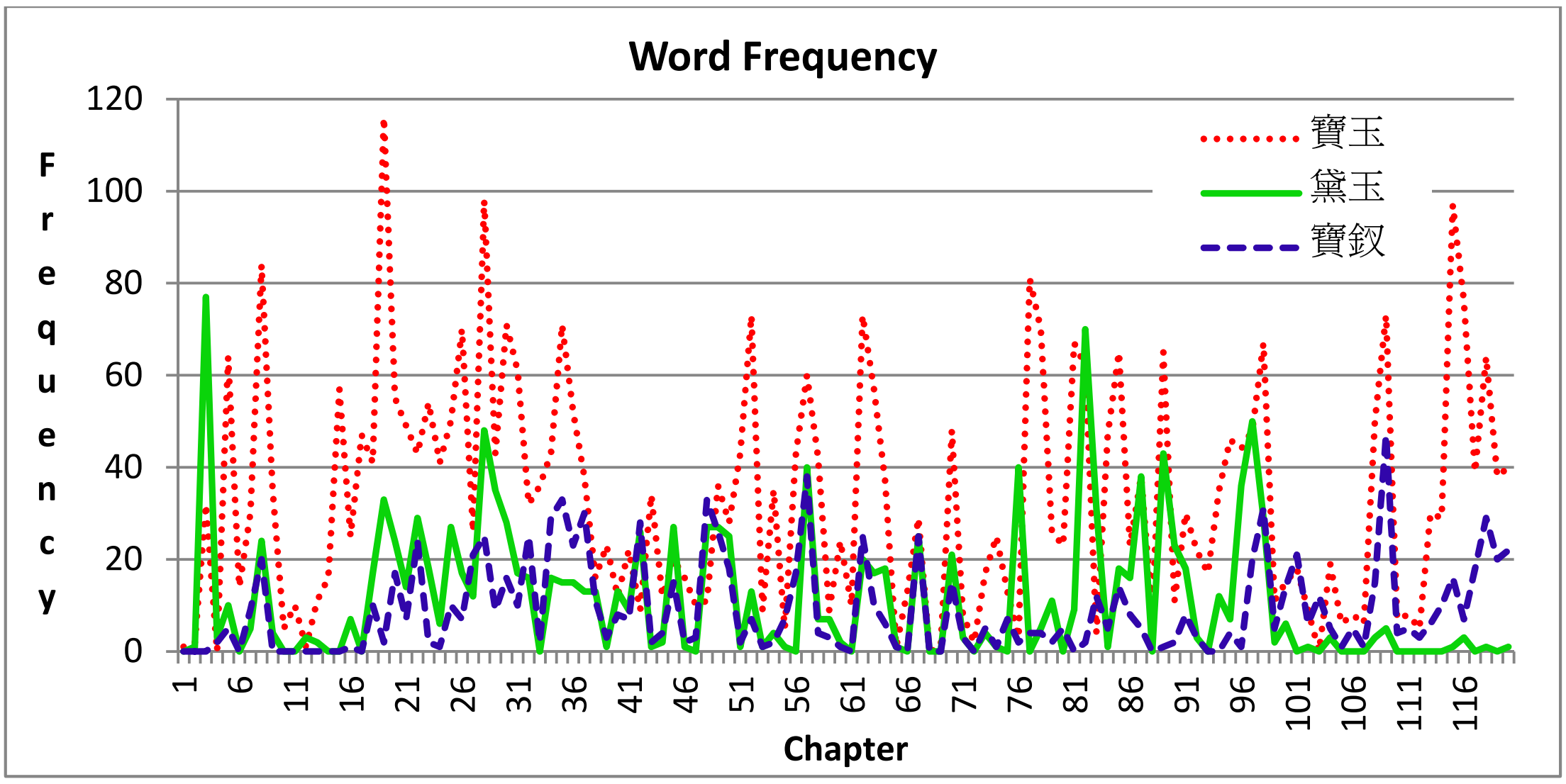

**Figure 5. Frequencies of three main names in *Dream of the Red Chamber***

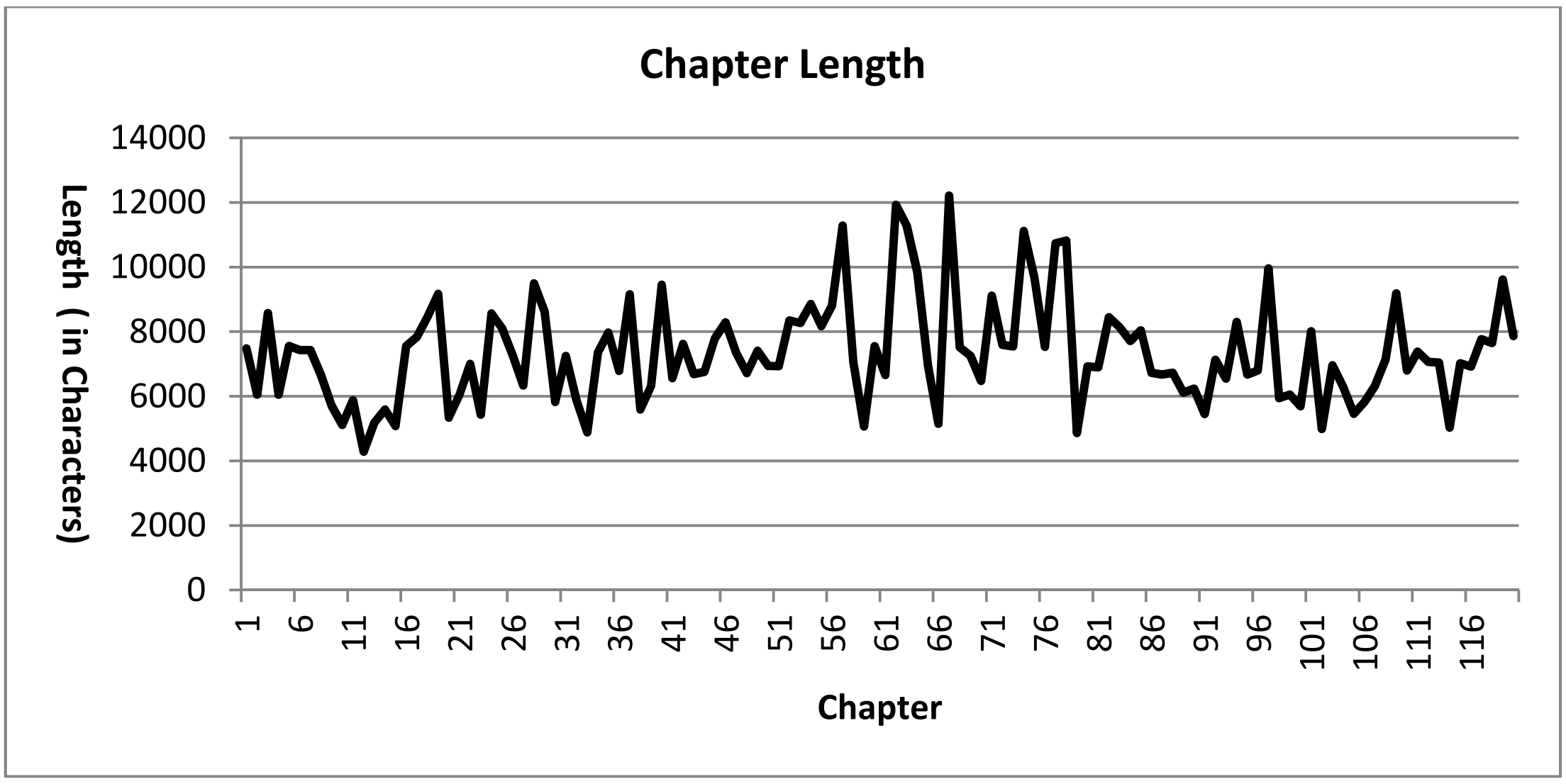

**Figure 6. Lengths of chapters in *Dream of the Red Chamber***

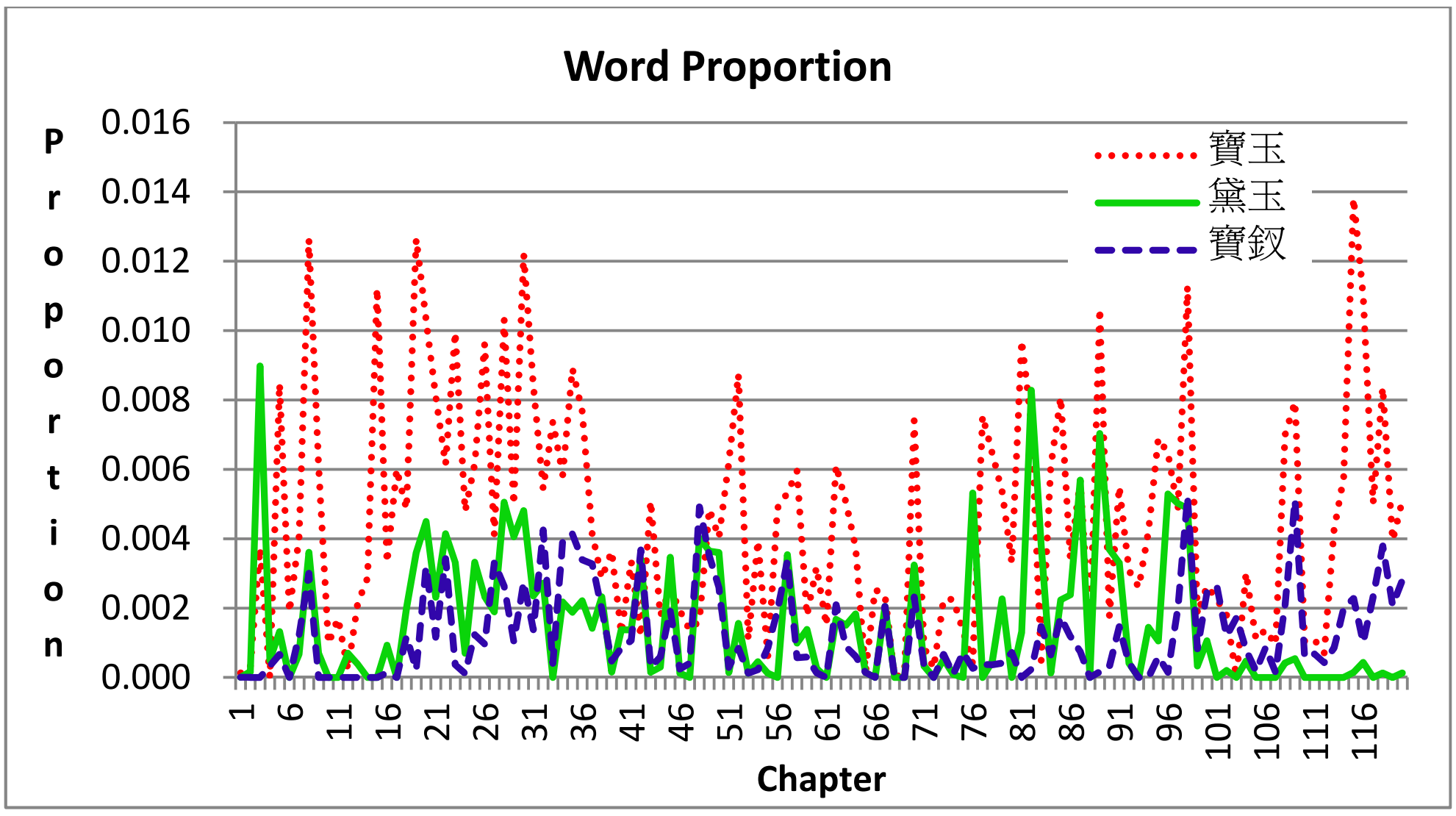

**Figure 7. Proportions of three major names in individual chapters in *Dream of the Red Chamber***

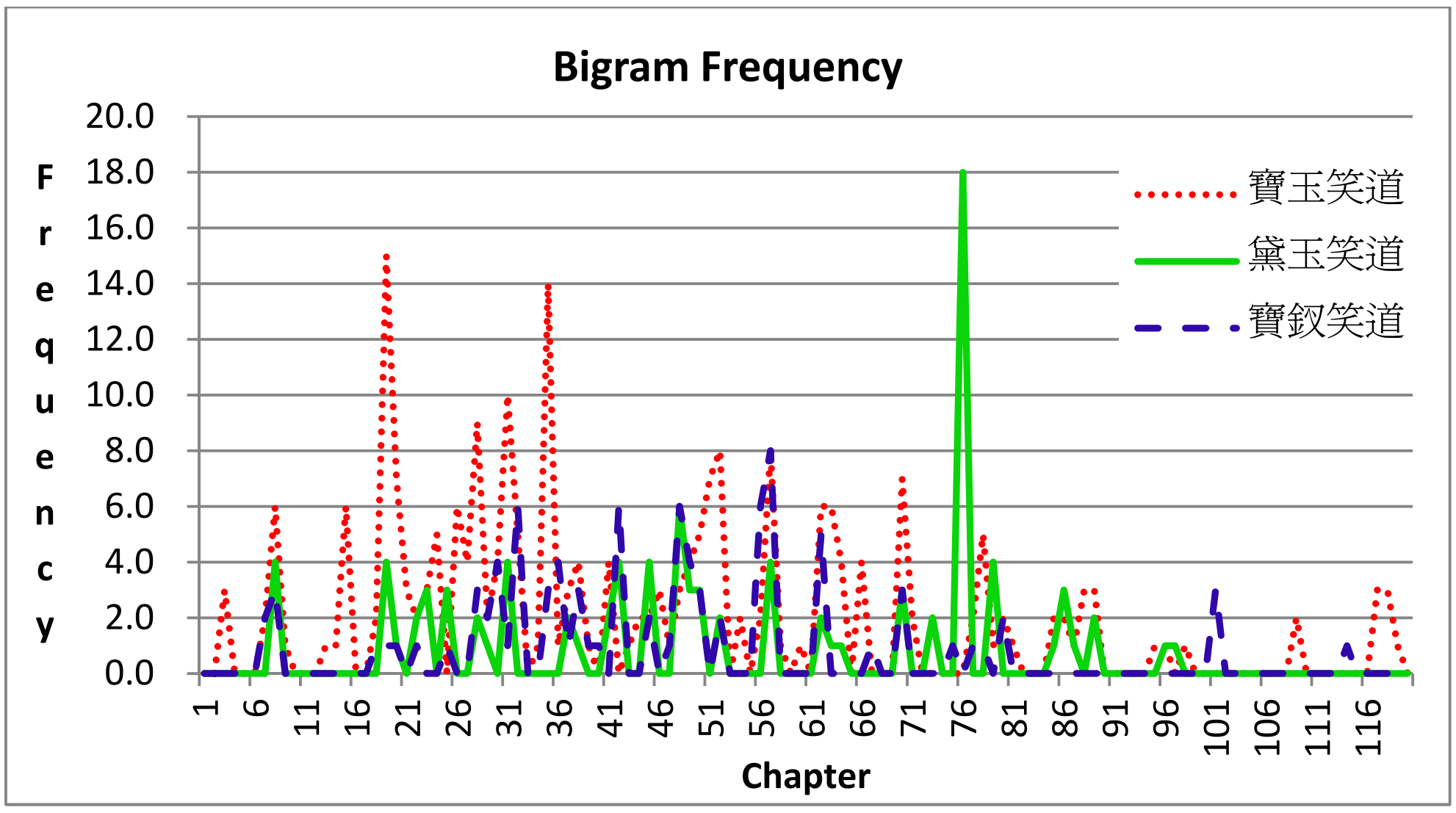

**Figure 8. Word frequencies indicate that Bow-Yu laughed most in early chapters**

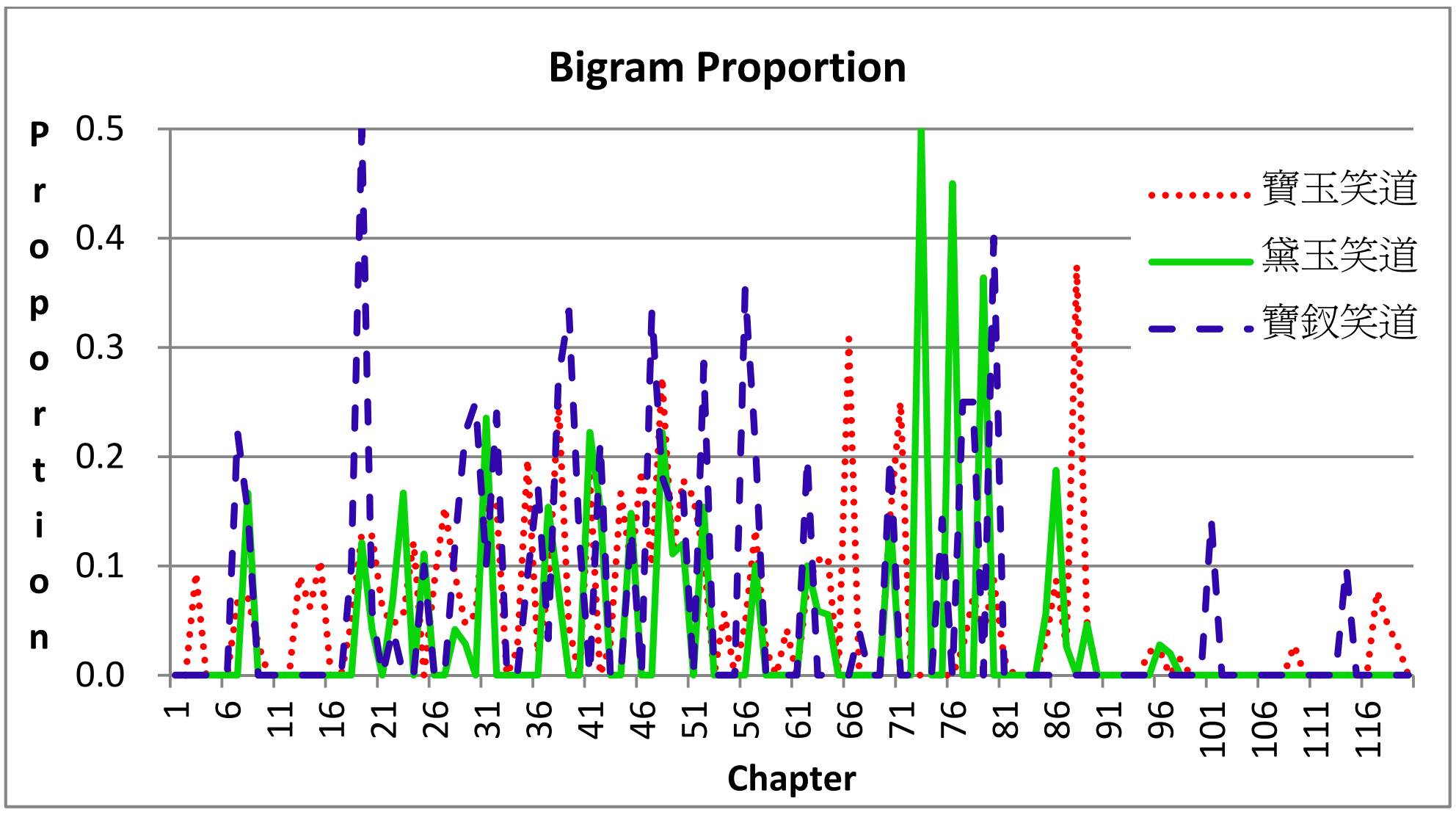

**Figure 9. Bigram proportions show that Bao-Chai laughed most in early chapters**